\documentclass[lettersize,journal]{IEEEtran}
\usepackage{amsmath,amsfonts}
\usepackage{algorithmic}
\usepackage{algorithm}
\usepackage{array}
\usepackage[caption=false,font=normalsize,labelfont=sf,textfont=sf]{subfig}
\usepackage{textcomp}
\usepackage{stfloats}
\usepackage{url}
\usepackage{verbatim}
\usepackage{graphicx}
\usepackage{cite}
\usepackage{tabularx}
\usepackage{multirow}
\usepackage{booktabs}
\usepackage{xcolor} 
\usepackage{colortbl}
\begin{document}

\title{Robust RGB-T Tracking via Learnable Visual Fourier Prompt Fine-tuning and Modality Fusion Prompt Generation}

\author{Hongtao Yang, Bineng Zhong\(^*\), Qihua Liang\(^*\), Zhiruo Zhu, Yaozong Zheng, Ning Li
\thanks{Hongtao Yang, Bineng Zhong, Qihua Liang, Zhiruo Zhu, Yaozong Zheng and Ning Li are with the Key Laboratory of Education Blockchain and Intelligent Technology, Ministry of Education, Guangxi Normal University, Guilin 541004, China, and the Guangxi Key Lab of Multi-Source Information Mining and Security, Guangxi Normal University, Guilin 541004, China.}
\thanks{Bineng Zhong and Qihua Liang are the corresponding authors.}}

\maketitle

\begin{abstract}
Recently, visual prompt tuning is introduced to RGB-Thermal (RGB-T) tracking as a parameter-efficient finetuning (PEFT) method. However, these PEFT-based RGB-T tracking methods typically rely solely on spatial domain information as prompts for feature extraction. As a result, they often fail to achieve optimal performance by overlooking the crucial role of frequency-domain information in prompt learning. To address this issue, we propose an efficient Visual Fourier Prompt Tracking (named VFPTrack) method to learn modality-related prompts via Fast Fourier Transform (FFT). Our method consists of symmetric feature extraction encoder with shared parameters, visual fourier prompts, and Modality Fusion Prompt Generator that generates bidirectional interaction prompts through multi-modal feature fusion. Specifically, we first use a frozen feature extraction encoder to extract RGB and thermal infrared (TIR) modality features. Then, we combine the visual prompts in the spatial domain with the frequency domain prompts obtained from the FFT, which allows for the full extraction and understanding of modality features from different domain information. Finally, unlike previous fusion methods, the modality fusion prompt generation module we use combines features from different modalities to generate a fused modality prompt. This modality prompt is interacted with each individual modality to fully enable feature interaction across different modalities. Extensive experiments conducted on three popular RGB-T tracking benchmarks show that our method demonstrates outstanding performance.

\end{abstract}


\section{Introduction}
\IEEEPARstart{R}{GB}-Thermal (RGB-T) tracking \cite{CAT,APFNet,DAPnet,DAFnet,tgtrack,tmm_3} is an important research area in visual tracking, where the objective is to estimate and localize the target's position in subsequent video frames based on the initial target state in the first frame. This is accomplished by integrating both RGB and thermal infrared images, which offer complementary information to enhance tracking accuracy under diverse environmental conditions. However, due to the differences between these two modalities, effectively extracting and fusing features from them remains an open problem.

\begin{figure}[!t]
\centering
\includegraphics[width=\linewidth]{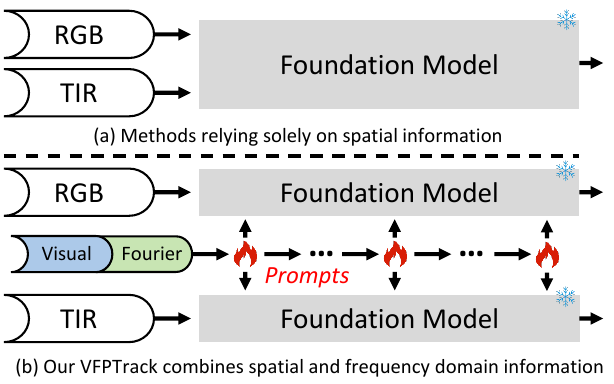}
\caption{\textbf{Methods relying solely on spatial information vs. VFPTrack.} (a) Previous methods that rely solely on spatial information focus only on spatial feature extraction, overlooking the positive impact of frequency domain information. (b) Our proposed VFPTrack introduces frequency domain prompt learning using Fast Fourier Transform (FFT), achieving fine-tuning of the backbone model by combining spatial and frequency domain information.}
\label{fig:one}
\end{figure}
The previous research \cite{CAT, APFNet, mfDiMP, SiamCDA} on RGB-T tracking mainly focus on designing complex fusion modules to fully utilize the complementary information from the two modalities. Recently, with the impressive success of existing RGB-based visual object tracking \cite{ostrack, odtrack, NingLi, artrack, tmm_1, tmm_2}, the attention of the research community has turned to fine-tuning frozen pre-trained RGB foundation models to adapt them for the RGB-T tracking task \cite{ViPT, BAT, sdstrack,untrack}, as shown in Fig. \ref{fig:one}(a). For example, ViPT \cite{ViPT} embeds lightweight modality-complementary prompters into frozen foundation model, achieving modality complementarity by using the spatial domain features of the auxiliary modality as prompts for learning. On the other hand, BAT \cite{BAT} introduces bidirectional adapters in each transformer block of the frozen foundation model, dynamically extracting effective features from the spatial domain through a non-fixed dominant-auxiliary modality, and adaptively fusing multi-modal features. However, the aforementioned methods focus solely on fine-tuning through spatial domain prompts, overlooking the importance of frequency-domain information, which results in suboptimal performance. As is well known, frequency-domain information is crucial for visual tasks \cite{10163861, Duhamel_Vetterli_1990, chu2023rethinking, oppenheim1979phase, Sangwine_Ell_2002, zeng2024visual} as it provides valuable insights for a model to understand image features. This naturally arises a question: Can we effectively leverage frequency information for any purpose?

Our answer is yes. In this paper, we propose a RGB-T tracker named VFPTrack, which leverages visual prompts to introduce spatial-domain information and Fourier prompts to incorporate frequency-domain information for fine-tuning the foundation model, as shown in Fig. \ref{fig:one}(b). Specifically, based on the characteristics of frequency-domain and spatial-domain information, we designed several lightweight visual Fourier prompt embeddings within the frozen ViT blocks. Visual Fourier prompts help integrate different domain information during the feature extraction stage, enabling more effective feature extraction. Additionally, the visual Fourier prompts from different layers are propagated to deeper layers, progressively capturing more complex features. Next, we use a simple modality fusion prompt generation network, which generates a simple yet effective fused modality prompt based on RGB modality features and TIR modality features. Finally, to enable the adaptive flow of features from different modalities and build complementary relationships, we ensure full interaction between the fusion modality prompt and features from different modalities at each layer. Extensive experiments on multiple RGB-T datasets demonstrate that our proposed VFPTrack outperforms existing RGB-T trackers and achieves robust performance in several challenging scenarios. Our main contributions are summarised as follows:

\begin{itemize}
    \item We propose a learnable visual Fourier prompt fine-tuning method for RGB-T tracking that integrates spatial and frequency domain information, enabling the frozen backbone network to adapt effectively. Additionally, an adaptive prompt transformation mechanism updates RGB and TIR prompts iteratively by incorporating information from the opposite modality at each Transformer layer, enhancing multi-modal integration and improving RGB-T tracking performance.
    \item To explore the fusion of different modalities, we propose a modality fusion prompt generation module. It generates a simple yet efficient fused modality prompt, enabling the features from various modalities to adaptively flow and build complementary relationships.
    \item Owing to the incorporation of frequency-domain information through Fast Fourier Transform and the design of the modality fusion prompt generation module, our VFPTrack achieves outstanding performance on three popular RGB-T tracking benchmarks, including LasHeR, RGBT210, and RGBT234.
\end{itemize}

\section{RELATED WORK}
\subsection{RGB-T Tracking}
Effectively fusing features from different modalities is crucial for enhancing RGB-T tracking performance. Similar challenges have also been investigated in RGB-T salient object detection, where position-aware relation learning \cite{a} and frequency-aware feature aggregation \cite{b} were proposed to mitigate cross-modal inconsistency. These works highlight that modality discrepancy is a fundamental obstacle across multi-modal vision tasks, underscoring its significance for RGB-T tracking as well. To address this issue, numerous methods \cite{CAT, APFNet, fanet,mfDiMP, c} have been actively explored. FANet \cite{fanet} employs a feature aggregation module for intra-modal feature fusion, while an adaptive aggregation module handles feature fusion across modalities, enhancing multi-modal information integration. APFNet \cite{APFNet} takes a different approach by designing dedicated attribute fusion branches for challenging scenarios and constructing an attribute-based progressive fusion network, offering new perspectives on modality fusion in complex environments. mfDiMP \cite{mfDiMP} systematically investigates three levels of multi-modal fusion—pixel, feature, and response levels-aiming to integrate multi-modal information efficiently in an end-to-end manner. However, a major drawback of these traditional methods is their inevitable introduction of redundant noise, leading to decreased cross-modal accuracy. With recent technological advancements, Transformer-based methods \cite{ViPT,tbsi,sdstrack,BAT} have emerged. For instance, TBSI \cite{tbsi} ingeniously uses templates as a crucial medium. By gathering and distributing target-related object and environmental information, it mitigates background noise interference and enhances fusion effectiveness. On the other hand, BAT \cite{BAT} utilizes the Transformer to extract features. It embeds a bidirectional adapter in each layer of the Transformer model, thereby enabling dynamic modality interaction. Through this approach, BAT achieves adaptive feature prompt fusion.

Different from the aforementioned methods, the feature fusion module we designed neither relies on complex modules nor is confined to the fusion of specific layers. Instead, on the premise of reducing redundant features, it conducts sufficient interactions at each layer. Moreover, with the incorporation of frequency-domain prompts and spatial-domain prompts, it becomes easier to distinguish between target and background features. This approach can enhance the accuracy and effectiveness of cross-modal fusion while minimizing redundant noise to the greatest extent. 

\subsection{Spatial and Frequency Domain Information}
Several studies \cite{low1, low2,low3,low4,low5} have revealed that the Transformer essentially acts as a low-pass filter. When processing images or sequential data, the self-attention mechanism prioritizes global, long-range dependencies, capturing primarily low-frequency information with a broad impact. However, it exhibits relatively weaker capabilities in capturing high-frequency details. In practical applications, data often contains a rich mix of low and high-frequency components, which complement each other to characterize the data more comprehensively. Introducing frequency domain information enables the model to analyze data at multiple frequency scales, facilitating a more in-depth understanding. SFTransT \cite{SFTransT} employs a spatio-frequency Transformer to simultaneously model spatial priors and frequency information, effectively preserving high-frequency features within stacked self-attention layers for robust tracking. Zhou et al. \cite{b} proposed a frequency-aware multi-frequency feature aggregation module for RGB-T salient object detection (SOD), which generates and fully utilizes both low- and high-frequency signals. SFDFusion \cite{sfdf} introduces a Frequency Domain Fusion Module (FDFM), leveraging the Fast Fourier Transform (FFT) to transform spatial domain features into the frequency domain before integration. 

However, these methods input the frequency information of the entire image into their designed modules, which may introduce redundant data and increase computational overhead. Zeng et al. \cite{zeng2024visual} incorporated FFT into prompt embeddings to integrate spatial and frequency domain information. However, their approach directly injects the same spatial and frequency domain prompts into every Transformer layer without hierarchical adaptation or cross-modal interaction. This static embedding strategy limits the model’s ability to refine and propagate multi-modal frequency cues across different representation levels. In contrast, our method introduces a hierarchical and interactive fusion mechanism. Instead of repeatedly injecting identical prompts at each layer, we enable the spatial and frequency domain prompts from different modalities to interact dynamically across hierarchical levels. As the Transformer layers deepen, the prompts are progressively updated based on the multi-modal representations from the previous layers, allowing information to flow and evolve through the encoder. This hierarchical propagation ensures that the model learns contextualized frequency-aware representations at different semantic levels, leading to a more expressive and adaptive integration of spatial and frequency domain features.

\begin{figure*}[!t]
\centering
\includegraphics[width=\linewidth]{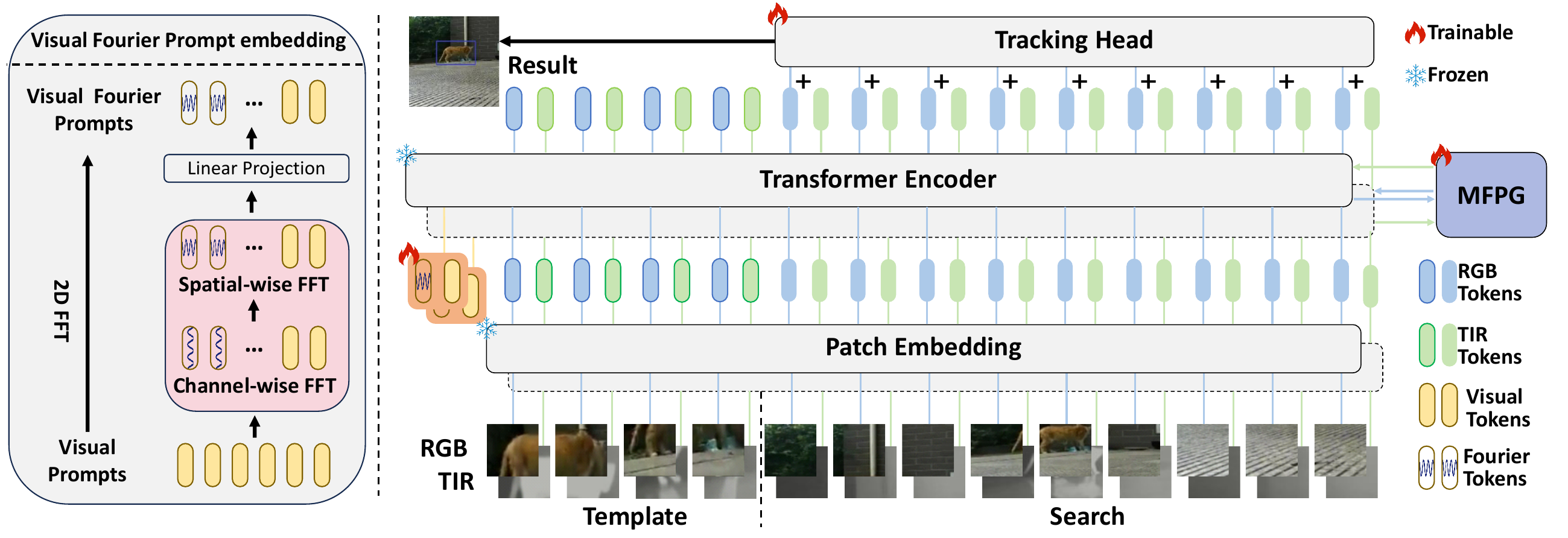}
\caption{\textbf{Overview of our proposed VFPTrack.} The RGB and TIR modality inputs are first fed into the patch embedding layer to generate corresponding tokens, which are then concatenated with the visual Fourier prompts. \( N \)-layer stacked transformer encoders are used to extract features from each modality separately. The MFPG module fuses the features of both modalities to achieve inter-modal complementarity. Finally, the RGB and TIR features are combined and forwarded to the tracking head network to obtain the prediction results. The structure of the visual Fourier prompt embedding is shown on the left.
}
\label{fig_1}
\end{figure*}

\begin{figure}[!t]
\centering
\includegraphics[width=\linewidth]{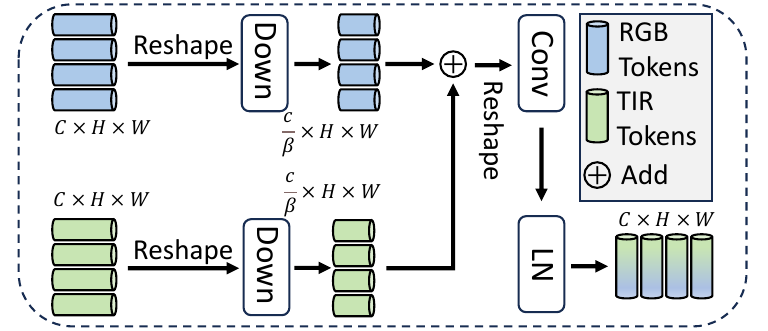}
\caption{\textbf{Detailed design of the proposed MFPG.} It consists of linear projection layer, convolutional layer, and normalization layer (LN). The input streams of the two modalities are first passed through the linear projection layer, where the channel dimension is reduced from \(C\) to \(\frac{C}{\beta}\), and then the features of the two modalities are combined by addition. Next, the features are fed into the convolutional layer for fusion and mapping back to the original dimension, and finally, the multi-modal prompts are obtained through the LN.}
\label{fig_3}
\end{figure}
\section{APPROACH}
In this section, we introduce a novel visual Fourier prompt framework for RGB-T tracking, named VFPTrack. First, we provide some preliminary knowledge and an initial introduction to the visual Fourier prompt fine-tuning approach. Then, we will explain how the Fourier prompts are generated and how they act as prompts during feature extraction. Specifically, we highlight the interaction between frequency-domain and spatial-domain prompts, where these prompts iteratively update and integrate information from the opposite modality at each layer, enhancing the model’s ability to capture complementary features. Additionally, the MPFG module and its complementary effects on RGB and TIR modalities are described in detail. Finally, we introduce the tracking head and loss function.

\subsection{Preliminary}
We proposed VFPTrack which enables the tracker to understand the target under different modalities by leveraging various domain information, such as the spatial and frequency domains. It also facilitates cross-modal interaction through a simple yet effective modality fusion module. The overall architecture is shown in Fig. 2. The encoder takes the template image Z, search image X, and visual 
Fourier prompt T as inputs, where the template frame and search frame are projected into feature space and flattened, allowing the use of a patch embedding layer to generate token embeddings:

\begin{flalign}
    \begin{split}
    &\mathbf{Z}\in\mathbb{R}^{H_{z}\times W_{z}\times3} \rightarrow \mathbf{Z}^{P}\in\mathbb{R}^{N_{z}\times D},\\
    &\mathbf{X}\in\mathbb{R}^{H_{x}\times W_{x}\times3} \rightarrow \mathbf{X}^{P}\in\mathbb{R}^{N_{x}\times D},\\
    \end{split}
\end{flalign}
where \(N_{z} = H_{z}\times W_{z} / P^{2}, N_{x} = H_{x}\times W_{x} / P^{2}, D = P^{2}\times C\) (D is the number of channel dimensions of the token embeddings generated using the patch embedding layer). 

We embed a set of d-dimensional embedding vectors (i.e., visual prompts) at each of the N layers of the pretrained encoder. These learnable prompts are defined as \( T = \{T_1, T_2, ..., T_N\} \), where \( T_i \) represents the learnable visual Fourier prompt embedded at the i-th encoder layer. These learnable visual Fourier prompts consist of two parts: the visual prompt of spatial domain information and the visual prompt of frequency domain information generated through Fast Fourier Transform, which can be defined as \( T_i = \{T^{v}_{i}; T^{f}_{i}\} \). The feature extraction process in the encoder, which differs from the foundation model, is as follows:
\begin{flalign}
    \begin{split}
    \mathbf{Encoder}_{i}&(Z_i;X_i)\\
    &\Downarrow\\
    \textcolor{blue}{\mathbf{Encoder}_{i}}&(\textcolor{red}{T_i};Z_i;X_i)\\
    \end{split}
\end{flalign}
where \( i \in \{1, 2, ..., N\} \), with the  \textcolor{blue}{blue} representing the frozen parameters and the  \textcolor{red}{red} representing the trainable parameters.

\subsection{Visual Fourier Prompt Embedding}
The Fast Fourier Transform (FFT) is a powerful algorithm used to compute the Discrete Fourier Transform (DFT). The DFT in image (i.e., two-dimensional space) is defined as:
\begin{equation}
    X[k_x, k_y] = \sum_{n_x=0}^{N_x-1} \sum_{n_y=0}^{N_y-1} x[n_x, n_y] \cdot e^{-j \frac{2\pi}{N_x} k_x n_x} \cdot e^{-j \frac{2\pi}{N_y} k_y n_y}
\end{equation}
where \( x[n_x, n_y] \) is the signal in the two-dimensional spatio-temporal domain, and \( X[k_x, k_y] \) is its corresponding frequency domain representation. 

We introduce learnable visual prompts to the input of each layer in the encoder. For the \(i\)-th layer, we represent the learnable visual prompts as \( T_i = \{ T^{k}_{i} \mid 1 \leq k \leq M \} \), where \(M\) is the number of learnable visual prompts chosen. Additionally, we transform the \(m\) visual prompts (where \(0 \leq m \leq M\)) from the spatial domain to the frequency domain, ultimately generating Fourier prompts, represented as \( TF_{i} = \{ TF^{k}_{i} \in T_i \mid 1 \leq k \leq m \} \). The conversion from spatial domain information to frequency domain information is primarily achieved through 2D Fast Fourier Transform (FFT). First, the 1D FFT is applied to the last dimension of the input tensor (i.e., the channel dimension), followed by the application of the 1D FFT to the second-to-last dimension (i.e., the spatial dimension). The operation process can be expressed by the following formula:
\begin{equation}
\begin{split}
T^{'} &= \mathcal{FFT}_c(T) = \sum_{k=0}^{c-1} T_{b, m, k} \cdot e^{-2\pi i \frac{uk}{c}}, \quad u \in [0, c-1] \\
T^{''} &= \mathcal{FFT}_m(T^{'}) = \sum_{k=0}^{m-1} T^{'}_{b, k, c} \cdot e^{-2\pi i \frac{vk}{m}}, \quad v \in [0, m-1] \\
TF &= \text{Re}(T^{''}) \\
TF &= \text{Re} \left( \mathcal{FFT}_m \left( \mathcal{FFT}_c(F) \right) \right) \\
\end{split}
\end{equation}
where \( FFT_m \) is the 1D Fourier transform performed along the \( m \)-th dimension (the spatial dimension),  \( FFT_c \) is the 1D Fourier transform performed along the \( C \)-th dimension (the channel dimension),  \( \text{Re} \) represents the magnitude obtained after applying the Fast Fourier Transform.

\subsection{Multi-modal Tracking and MFPG}
Multi-modal tracking provides two different modalities of input, which are synchronized in time and aligned in space. The RGB and TIR modal inputs are fed into the patch embedding layer separately, which generate the corresponding token embeddings. These embeddings are then concatenated with visual Fourier prompts, and passed to the foundation model encoder to extract features for the two modalities separately. The features extracted by each layer of the encoder are represented as \( F^{i}_{\text{rgb}} \) and \( F^{i}_{\text{tir}} \) for the RGB and TIR modalities, where  \( F^{i}_{\text{rgb}} \) = \{\( Z^{i}_{\text{rgb}} \);  \( X^{i}_{\text{rgb}} \)\},  \( F^{i}_{\text{tir}} \) = \{\( Z^{i}_{\text{tir}} \);  \( X^{i}_{\text{tir}} \)\} and i represents the \(i\)-th encoder layer. Then, we input the features of the two different modalities into the modality fusion prompt generator (MFPG). By combining the features of both modalities, a modality prompt is generated. The modality prompt will be added to the different modalities in the form of a residual:
\begin{equation}
\label{eq2}
	\left\{
	\begin{aligned}
        {F}^{i}_{\text{rgb}} &= {F}^{i}_{\text{rgb}} + {P}^{i}\\
        {F}^{i}_{\text{tir}} &= {F}^{i}_{\text{tir}} + {P}^{i}
	\end{aligned}
	\right.
\end{equation}
where \({P}^{i}\) represents the prompt token generated by the i-th layer modal fusion prompt generator. We will embed the MFPG module in each feature extraction block to fully integrate the features extracted from different modalities at each level. It is important to emphasize that we follow previous practices \cite{ViPT}\cite{BAT} by freezing the relevant parameters of the patch embedding and the Transformer encoder.

\textbf{Interaction Between Spatial and Frequency Domain Prompts.}
To enhance hierarchical and cross-modal fusion, an adaptive prompt transformation mechanism is introduced. At each Transformer layer, the RGB and TIR prompts are dynamically updated by incorporating the previous-layer prompt from the opposite modality. This interaction is realized through a learnable linear projection, which refines and transforms the incoming prompt features before they are integrated into the current-layer representation. Starting from the second layer, the prompt tokens are no longer static but progressively refined through residual fusion with the transformed counterpart prompts, ensuring that complementary cues from both modalities are adaptively propagated across layers. Formally, the update can be expressed as:

\begin{equation}
T^l = T_{\mathrm{init}}^l + \mathcal{F}\left(TF^{l-1}\right),
\end{equation}

where $T_{\mathrm{init}}^l$ denotes the initialized RGB prompt tokens at layer $l$, $TF^{l-1}$ denotes the TIR prompts from the previous layer, and $\mathcal{F}(\cdot)$ represents the learnable transformation module. This update enables the RGB prompts to evolve hierarchically under prior TIR guidance. Similarly, in the TIR branch, the initialized TIR prompts are refined by incorporating the transformed RGB prompts from the previous layer. Through such symmetric updates, the prompts progressively capture complementary multi-modal information, leading to more expressive cross-modal representations.

\textbf{Modality Fusion Prompt Generator.}
The detailed design of MFPG is shown in Fig. \ref{fig_3}. It takes inputs from two modalities, fuses the features of both modalities to achieve complementarity, and then incorporates the fused features as prompts into both modalities. However, there is redundancy in the modal features, and some unnecessary features need to be filtered out, which also helps capture more effective modal features. Therefore, we adaptively reduce the channel dimensions through projection layer: 
\begin{flalign}
    \begin{split}
    {PF}^{i}_{\text{rgb}} = \mathcal{P}roj_1({F}^{i}_{\text{rgb}}), \quad {PF}^{i}_{\text{tir}} = \mathcal{P}roj_1({F}^{i}_{\text{tir}}),\\
    \end{split}
\end{flalign}
where \(\mathcal{P}roj_1\) is responsible for reducing the channel dimension of the modal features from \(C\) to \(\frac{C}{\beta}\). In our implementation, \(\mathcal{P}roj_1\) is a simple linear layer, and \(\beta\) is set to 96, meaning that the input with 768 channels is projected to a channel dimension of 8. Then, since the two modalities correspond spatially, we choose to combine them by simple addition and apply 3 \(\times\) 3 convolutional layer to perform the fusion operation. During the fusion process, we project the resulting features back to the original dimension to generate the learned multi-modal visual prompts:
\begin{flalign}
    \begin{split}
    {P}^{i} = \mathcal{C}^{\uparrow}({PF}^{i}_{\text{rgb}} + {PF}^{i}_{\text{tir}}), \\
    \end{split}
\end{flalign}
Finally, we normalize the obtained multi-modal visual prompt and add it to the two modalities through a residual connection to enhance the representation ability.

\subsection{Prediction Head and Loss Function}
The encoder will ultimately output features from both modalities. Before feeding them into the prediction head, we mix the features of the two modalities using an additive binding. The design of the prediction head still relies on a center-point-based prediction head to predict the target's center location and scale. By converting the mixed modality embedding from 1D feature maps to 2D spatial features, they are then passed into a fully convolutional network to obtain the final tracking results. We use the focal \cite{focal} loss as the classification loss, the GIoU \cite{giou} loss, and the L1 loss as the regression loss, whose total loss can be expressed as:
\begin{equation}
    \begin{aligned}
    {L} &= {L}_{cls} + {\lambda}_{Giou}{L}_{Giou} + {\lambda}_{L_{1}}{L}_{1},\\
    \end{aligned}
\end{equation}
where ${\lambda}_{L_{1}}$ = 5 and ${\lambda}_{Giou}$ = 2 are the regularization parameters.

\section{Experiments}
\begin{table*}[t]
    \centering
    \small 
    \setlength{\tabcolsep}{6.4pt} 
    \renewcommand{\arraystretch}{1.3} 
    \caption{Performance Comparisons with state-of-the-art trackers on the test set of rgbt210 \cite{rgbt210}, rgbt234 \cite{rgbt234}, and lasher \cite{lasher}. The top two results are highlighted with red and blue fonts, respectively.}
    \begin{tabular*}{\linewidth}{l|c|c|cc|cc|ccc|c|c}
        \toprule
        \multirow{2}{*}{\textbf{Method}} & \multirow{2}{*}{\textbf{Source}} & \multirow{2}{*}{\textbf{Backbone}} & \multicolumn{2}{c|}{\textbf{RGBT210}} & \multicolumn{2}{c|}{\textbf{RGBT234}} & \multicolumn{3}{c|}{\textbf{LasHeR}} & \multirow{2}{*}{\textbf{FPS}} & \multirow{2}{*}{\textbf{Device}}\\
        & & &  SR \(\uparrow\) & PR \(\uparrow\) &  MSR \(\uparrow\) & MPR \(\uparrow\) &  SR \(\uparrow\) & PR \(\uparrow\) & NPR \(\uparrow\) &\\
        \midrule
        mfDiMP\cite{mfDiMP} & ICCVW-19 & ResNet-50 & 55.5 & 78.6 & 42.8 & 64.6 & 34.3 & 44.7 & 39.5 & 10.3 & - \\
        HMFT\cite{HMFT} & CVPR-22 & ResNet-50 & 53.5 & 78.6 & 56.8 & 78.8 & - & - & - & 30.2 & RTX Titan\\
        SiamMLAA\cite{TMM24_rgbt} & TMM-24 & ResNet-50 & 56.7 & 77.9 & 58.4 & 79.5 & 43.1 & 53.8 & - & 21.7 & GTX 1080Ti\\
        
        DAPNet\cite{DAPnet} & MM-19 & VGG-M & - & - & 53.7 & 76.6 & 31.4 & 43.1 & 38.3 & - & -\\
        MANet\cite{MANet} & ICCVW-19 & VGG-M & - & - & 53.9 & 77.7 & - & - & - & 1.1 & GTX 1080\\
        DAFNet\cite{DAFnet} & ICCVW-19 & VGG-M & - & - & 54.4 & 79.6 & 31.1 & 44.8 & 39.0 & 23 & RTX 2080Ti\\
        FANet\cite{fanet} & TIV-20 & VGG-M & - & - & 55.3 & 78.7 & 30.9 & 44.1 & 38.4 & 19.0 & GTX 2080Ti\\
        CAT\cite{CAT} & ECCV-20 & VGG-M & 53.3 & 79.2 & 56.1 & 80.4 & 31.4 & 45.0 & 39.5 & 20 & RTX 2080Ti\\
        CBPNet\cite{TMM22_RGBT} & TMM-22 & VGG-M & - & - & 54.1 & 79.4 & - & - & - & - & -\\
        DMCNet\cite{dmcnet} & TNNLS-22 & VGG-M & 55.5 & 79.7 & 59.3 & 83.9 & 35.5 & 49.0 & 43.1 & 2.3 & RTX 2080Ti\\
        APFNet\cite{APFNet} & AAAI-22 & VGG-M & - & - & 57.9 & 82.7 & 36.2 & 50.0 & 43.9 & 1.3 & GTX 1080Ti\\
        
        ViPT\cite{ViPT} & CVPR-23 & ViT-Base & - & - & 61.7 & 83.5 & 52.5 & 65.1 & - & 48.2 & Tesla v100\\     
        TBSI\cite{tbsi} & CVPR-23 & ViT-Base & 61.8 & 85.3 & 63.7 & 87.1 & 55.6 & 69.2 & 65.7 & 36.2 & RTX 3080Ti\\
        STMT\cite{STMT} & TCSVT-24 & ViT-Base & 59.5 & \textcolor{blue}{86.0} & 63.8 & 86.5 & 53.7 & 67.4 & 63.4 & 39.1 & RTX 2080Ti\\
        SDSTrack\cite{sdstrack} & CVPR-24 & ViT-Base & - & - & 62.5 & 84.8 & 53.1 & 66.5 & - & 20.8 & RTX 3090Ti\\
        TATrack\cite{tatrack} & AAAI-24 & ViT-Base & \textcolor{red}{62.5} & 85.3 & \textcolor{blue}{64.4} & \textcolor{blue}{87.2} & 56.1 & \textcolor{blue}{70.2} & \textcolor{blue}{66.7} & 26.1 & RTX 3090\\
        BAT\cite{BAT} & AAAI-24 & ViT-Base & - & - & 64.1 & 86.8 & \textcolor{blue}{56.3} & \textcolor{blue}{70.2} & - & 49.9 & Tesla v100 \\
        \midrule
        VFPTrack & Ours & ViT-Base & \textcolor{blue}{62.0} & \textcolor{red}{88.2} & \textcolor{red}{65.0} & \textcolor{red}{89.0} & \textcolor{red}{58.5} & \textcolor{red}{73.5} & \textcolor{red}{69.8} & 44.4 & Tesla v100 \\
        \bottomrule
    \end{tabular*}
    \label{table:two}
\end{table*}
In this section, we first describe the experimental implementation details of the proposed VFPTrack. Next, we will compare VFPTrack with the current SOTA trackers on three RGB-T tracking benchmarks including RGBT210 \cite{rgbt210}, RGBT234 \cite{rgbt234}, and LasHeR \cite{lasher}. Finally, we will conduct a performance analysis and ablation study on our tracker using the LasHeR \cite{lasher} dataset. Default settings are marked in \colorbox{gray!30}{\textbf{gray}}.
\subsection{Implementation Details}
\textbf{Model.} We initialize our model parameters using the pre-trained RGB-based model OSTrack \cite{ostrack} and freeze its backbone. For the visual Fourier prompts embedded in the transformer block, their channel dimensions match the channel dimensions of the image features, both being 768. Additionally, the template frame size for our proposed VFPTrack is [128 \(\times\) 128], and the search frame size is [256 \(\times\) 256].

\textbf{Training strategy.} Similar to previous approaches, we implement VFPTrack based on PyTorch and train it on four NVIDIA A800 GPUs with a batch size of 16, using only the LasHeR dataset for training. We follow the hyperparameter settings of the foundation model for the loss function, also using the AdamW \cite{AdamW} optimizer with weight decay set to \( 10^{-4} \) and a learning rate of \( 4 \times 10^{-4} \). The total training epochs are set to 60 with 60k image pairs per epoch, and we decrease the learning rate by a factor of 10 after 45 epochs.

\begin{table}[h]
    \centering
    \caption{Comparison of model parameters, FLOPs, and inference on the LasHeR \cite{lasher} dataset.}
    \resizebox{\linewidth}{!}{ 
    \begin{tabular}{l|c|c|c|c|c}
    \midrule
    \textbf{Model} & \textbf{Device} & \textbf{Speed(FPS)} & \textbf{MACs(G)} & \textbf{Params(M)} & \textbf{SR(\%)}\\
        \midrule 
        VFPTrack & Tesla v100 & 44.4 & 71.1 & 103.9 & 58.5\\
        \midrule
        BAT\cite{BAT} & Tesla v100& 49.9 & 56.6 & 92.4 & 56.3\\
        SDSTrack\cite{sdstrack} & Tesla v100& 17.9 & 108.3 & 102.1 & 53.1\\
        ViPT\cite{ViPT} & Tesla v100 & 48.2 & 21.8 & 92.9 & 52.5\\
        \bottomrule
    \end{tabular}
    }
    \label{table:one}
\end{table}

\textbf{Inference.} We follow the common practice \cite{Transt, ostrack, odtrack} of applying the Hanning window penalty to suppress displacement between inference frames. Additionally, as shown in Tab. \ref{table:one}, the proposed VFPTrack is tested on a single NVIDIA Tesla V100 GPU and it runs at 44.4 fps.

\subsection{Comparison With SOTA Trackers}
As shown in Tab. \ref{table:two}, we evaluated our proposed VFPTrack on three popular RGB-T tracking benchmarks: RGBT210 \cite{rgbt210}, RGBT234 \cite{rgbt234}, and LasHeR \cite{lasher}, and compared its performance with that of the state-of-the-art RGB-T trackers.

\textbf{RGBT210.} RGBT210 \cite{rgbt210} is a large-scale RGB-T tracking dataset for performance evaluation. The dataset includes 210 pairs of RGB-T videos, with each pair consisting of an RGB video and a TIR video. The total number of frames is 210k, with up to 8k frames per pair. The video pairs are highly aligned in both spatial and temporal dimensions. Additionally, each video sequence is annotated with occlusion attributes, categorized into no occlusion, partial occlusion, and total occlusion, enabling analysis of occlusion-sensitive performance across different methods. Success Rate (SR) and Precision Rate (PR) are used to evaluate the target performance.

\begin{figure}[!t]
    \centering
    \includegraphics[width=1\linewidth]{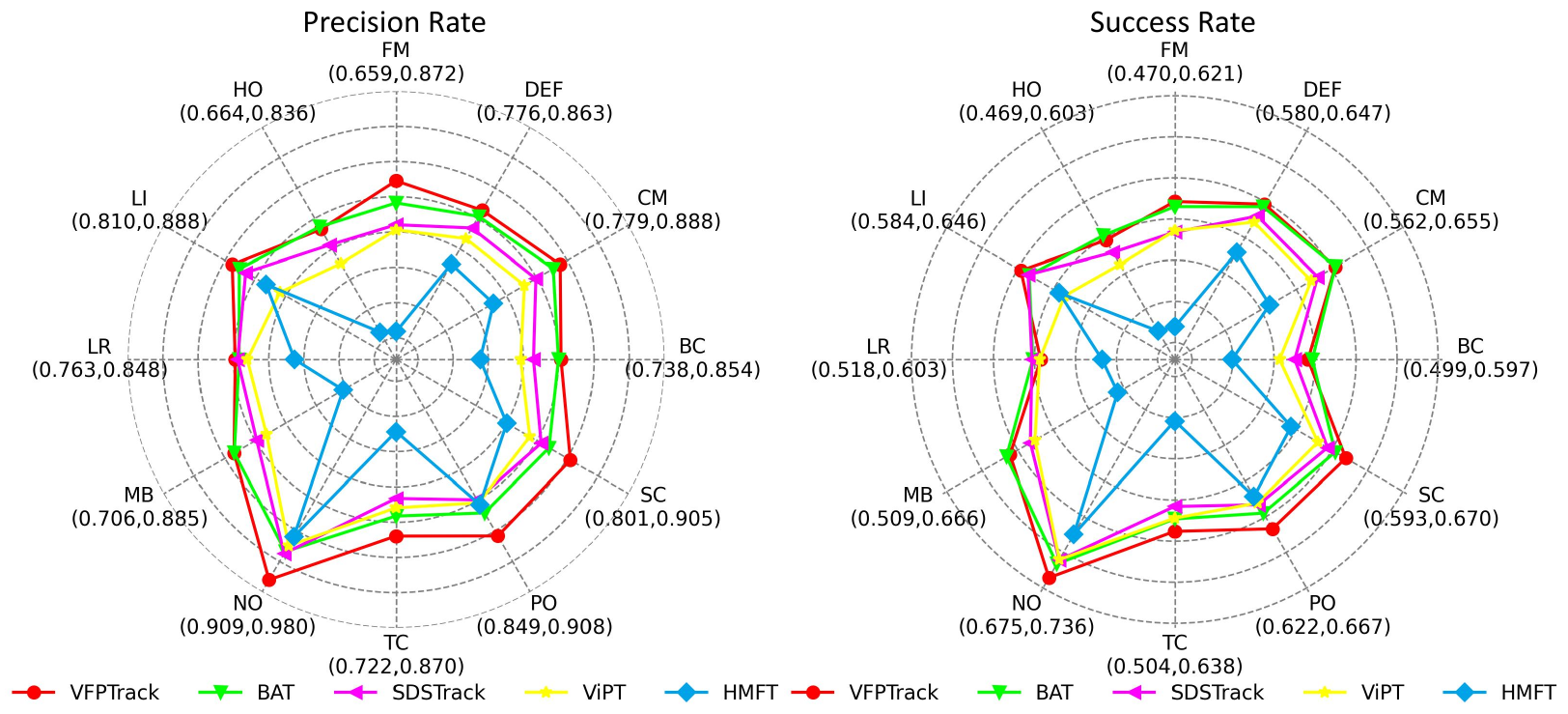}
    \caption{\textbf{More comparisons} of VFPTrack and the competing methods under different attributes in the RGBT234 \cite{rgbt234} dataset.}
    \label{fig:four}
\end{figure} 

\textbf{RGBT234.} The RGBT234 \cite{rgbt234} dataset is one of the most influential and widely evaluated RGB-T tracking datasets, consisting of 234 pairs of RGB and TIR videos, with a total of approximately 234K frames, and the longest video sequence having 8K frames. Performance is evaluated using Maximum Success Rate (MSR) and Maximum Precision Rate (MPR). We perform further comparisons between VFPTrack and competing methods under different attributes, with the results shown in Fig. \ref{fig:four}. The proposed method achieves the highest performance in terms of MSR and MPR, with values of 65.0\% and 89.0\%, respectively. Notably, it surpasses the previous best method by 1.8\% in MPR.

\textbf{LasHeR.} Compared to the previous RGB-T tracking benchmark, the LasHeR \cite{lasher} dataset is a large-scale, highly diverse benchmark for short-term RGB-T tracking. It contains 1,224 pairs of visible and thermal infrared videos, totaling over 730,000 frame pairs. Each frame pair is spatially aligned and annotated with manually labeled bounding boxes, providing the dataset with high-quality and dense annotations. It is diverse, encompassing various object categories, camera viewpoints, scene complexities, and environmental factors across different seasons, weather conditions, and times of day and night. We evaluate the trackers on 245 test video sequences, and the results are shown in Tab. \ref{table:two}. To the best of our knowledge, our method improves upon the existing state-of-the-art performance by 2.2\% in SR, 3.3\% in PR, and 3.1\% in NPR. This demonstrates that our proposed method achieves robustness in complex scenes and environmental factors by effectively utilizing thermal infrared information.

\begin{table}[h]
    \centering
    \caption{Ablation Study on the effect of components and modality inputs, along with performance comparison in challenging conditions. }
    \resizebox{\linewidth}{!}{ 
    \begin{tabular}{l|c|ccc}
    \midrule
    \multirow{2}{*}{\textbf{\#}} & \multirow{2}{*}{\textbf{Method}} & \multicolumn{3}{c}{\textbf{LasHeR}}\\
    & & Success \(\uparrow\) & Precision \(\uparrow\) & NormPrec \(\uparrow\)\\
    \midrule 
    1 & baseline & 52.5 & 66.9 & 62.7 \\
    \midrule
    2 & w/o RGB & 39.7 & 50.7 & 46.1 \\
    3 & w/o TIR & 50.9 & 64.0 & 59.6 \\
    \midrule
    4 & +MFPG & 56.0 & 70.4 & 67.0 \\
    \rowcolor{gray!30}
    \textbf{5} & \textbf{+Visual Fourier Prompt} & \textbf{58.5} & \textbf{73.5} & \textbf{69.8} \\
    \bottomrule
    \end{tabular}
    }
    \label{table:three}
\end{table}
\subsection{Ablation Studies}
\textbf{Component Analysis.} In Tab. \ref{table:three}, we conduct an ablation study on our proposed VFPTrack using the LasHeR dataset to analyze the effectiveness of different inputs and design components. The w/o RGB (Tab. \ref{table:three} \#2) represents the tracking head prediction network without the RGB features, while the w/o TIR (Tab. \ref{table:three} \#3) represents the network without the TIR features. The results indicate that combining the complementary features of both modal inputs effectively alleviates the limitations of a single modal input, even without separate training for the TIR modal input.

Incorporating the MFPG module into the baseline allows for a more comprehensive fusion of the two modality features, resulting in an impressive success rate (SR) of 56.0\%. Furthermore, the addition of the visual Fourier prompt leads to a further 2.5\% improvement in SR over the previous result. These findings clearly demonstrate the effectiveness of our proposed strategy and emphasize the role of visual Fourier prompts in improving the model's ability to extract features from different domain information.

\begin{table}[h]
    \centering
    \caption{Ablation Experiments on the MFPG Block of VFPTrack on the LasHeR  \cite{lasher} dataset.}
    \resizebox{\linewidth}{!}{ 
    \begin{tabular}{l|c|ccc}
    \midrule
    \multirow{2}{*}{\textbf{\#}} & \multirow{2}{*}{\textbf{layers}} & \multicolumn{3}{c}{\textbf{LasHeR}}\\
    & & Success \(\uparrow\) & Precision \(\uparrow\) & NormPrec \(\uparrow\)\\
    \midrule 
    1 & 1-6 & 58.1 & 72.7 & 69.1 \\
    2 & 7-12 & 58.4 & 73.1 & 69.6 \\
    \rowcolor{gray!30}
    \textbf{3} & \textbf{1-12} & \textbf{58.5} & \textbf{73.5} & \textbf{69.8} \\
    \bottomrule
    \end{tabular}
    }
    \label{table:four}
\end{table}

\begin{figure*}[!t]
    \centering
    \includegraphics[width=1\linewidth]{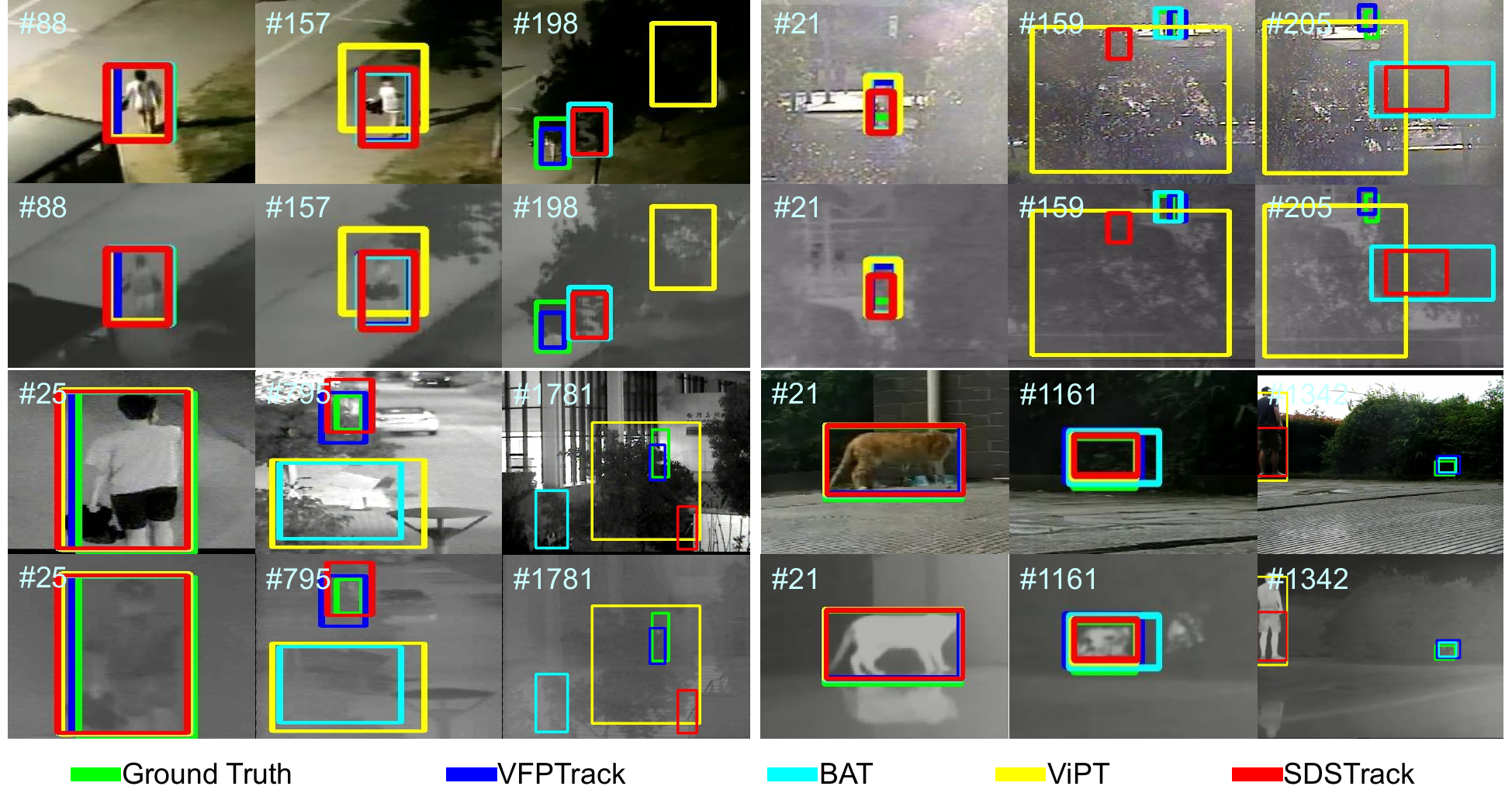}
    \caption{\textbf{Qualitative comparison} results of the proposed VFPTrack with other state-of-the-art trackers (e.g., BAT \cite{BAT}, ViPT \cite{ViPT} and SDSTrack \cite{sdstrack}) are evaluated on four challenging sequences from the LasHeR \cite{lasher} dataset. Best viewed in color.}
    \label{fig:seven}
\end{figure*}
\textbf{Number Analysis of MFPG Block.} 
VFPTrack attains the flow of multi-modal information by inserting MFPG blocks into diverse transformer block layers of the foundation model. The choice of where to insert these blocks is crucial. We conduct ablation experiments to explore the impacts of embedding MFPG blocks at different layers. The reason we first compare adding MFPG blocks to the first 6 layers and the last 6 layers of the feature extraction network is that the initial layers of a network typically capture low-level features, such as edges and textures, while the later layers are more concerned with high-level semantic features. By inserting MFPG blocks in the first 6 layers, we aim to see if the early fusion of multi-modal information can enhance the extraction of basic features. Inserting them in the last 6 layers might allow for better integration of semantic-level information from different modalities. Subsequently, we increase the number of embeddings by inserting MFPG blocks into every layer. As presented in Table \ref{table:four}, inserting MFPG blocks into each layer permits the features from both modalities to fully interact. This comprehensive interaction leads to a more thorough exchange of information between modalities. The model can then leverage the complementary nature of different modal data more effectively, resulting in improved performance. For example, visual and thermal modalities can share information at all levels of feature extraction, enabling the model to better distinguish objects in complex scenarios.

\begin{table}[h]
    \centering
    \caption{Ablation Studies of Our Fourier Prompt Dimension.}
    \resizebox{\linewidth}{!}{
    \begin{tabular}{l|c|c|ccc}
    \midrule
    \multirow{2}{*}{\textbf{\#}} & \multirow{2}{*}{\textbf{Spatial-wise}} & \multirow{2}{*}{\textbf{Channel-wise}} &\multicolumn{3}{c}{\textbf{LasHeR}}\\
    & & & Success \(\uparrow\) & Precision \(\uparrow\) & NormPrec \(\uparrow\)\\
    \midrule 
    1 & \(\checkmark\) & & 57.7 & 72.4 & 68.7 \\
    2 & & \(\checkmark\) & 58.4 & 73.1 & 69.6 \\
    \rowcolor{gray!30}
    \textbf{3} & \textbf{\(\checkmark\)} & \textbf{\(\checkmark\)} & \textbf{58.5} & \textbf{73.5} & \textbf{69.8} \\
    \bottomrule
    \end{tabular}
    }
    \label{table:five}
\end{table}

\textbf{Analysis of Different Dimensions in 2D FFT.} 
We apply 2D FFT to transform visual prompts into frequency domain prompts, first conducting a Fourier transform along the channel dimension and then along the spatial dimension. The reason for performing the Fourier transform along the channel dimension first is that different channels in an image often represent different features. Transforming along this dimension helps in analyzing the frequency characteristics of these feature channels. After that, the spatial dimension transform captures the spatial frequency distribution of the image.
Here, we discuss the different impacts of transformations in each dimension on performance improvement, with the results shown in Tab. \ref{table:five}. The Fourier prompts obtained by combining transformations in both dimensions enable synergistic prompting in the frequency domain. This is because the combination allows the model to take into account both the frequency information related to feature channels and the spatial frequency distribution simultaneously. By doing so, the model can better understand the global and local characteristics of the visual data in the frequency domain. As a result, it can generate more accurate and effective prompts, leading to the best performance. For instance, in object tracking, this can help the model better distinguish the target object from the background based on both the feature-specific and spatial-specific frequency information. 

\begin{table}[h]
    \centering
    \caption{Ablation experiments of the visual Fourier prompt layers on the LasHeR \cite{lasher} dataset.}
    \resizebox{\linewidth}{!}{
    \begin{tabular}{l|c|ccc}
    \midrule
    \multirow{2}{*}{\textbf{\#}} & \multirow{2}{*}{\textbf{layers}} & \multicolumn{3}{c}{\textbf{LasHeR}}\\
    & & Success \(\uparrow\) & Precision \(\uparrow\) & NormPrec \(\uparrow\)\\
    \midrule 
    1 & 1-6 & 57.9 & 72.6 & 69.3 \\
    2 & 7-12 & 57.6 & 72.4 & 68.8 \\
    \rowcolor{gray!30}
    \textbf{3} & \textbf{1-12} & \textbf{58.5} & \textbf{73.5} & \textbf{69.8} \\
    \bottomrule
    \end{tabular}
    }
    \label{table:six}
\end{table}

\textbf{Number Analysis of Visual Fourier Prompt Block.} 
Table \ref{table:six} demonstrates the performance with visual Fourier prompts at different layers. The reason we tested the use of visual Fourier prompts at different layers was to determine the optimal layer for integrating this type of prompt. Using visual Fourier prompts in some layers can improve performance because it allows the model to incorporate frequency-domain information at specific levels of feature extraction. For example, in the middle layers, where the features start to have a certain degree of semantic meaning but still retain some local details, the visual Fourier prompts can enhance the model's ability to capture complex visual patterns. Applying them to all layers achieves the best results as it provides a comprehensive integration of frequency-domain information throughout the entire feature extraction process. This way, the model can make full use of the additional information from the frequency domain at every stage of feature representation, leading to more accurate and robust performance.

\begin{table}[h]
    \centering
    \caption{OCCURRENCE RATIO OF $\alpha=20\%$ IN DIFFERENT ATTRIBUTES WITH DYNAMIC ADJUSTMENT.}
    \resizebox{\linewidth}{!}{
    \begin{tabular}{l|c|c|c}
    \midrule
    \multirow{2}{*}{\textbf{Sequences}} & \multirow{2}{*}{\textbf{Attributes}} & \multicolumn{2}{c}{\textbf{Occurrence Ratio (\%)}}\\
    & & RGB & TIR\\
        \midrule 
        10-roune & PO, BC & 89.36 & 100\\
        baggirl & FM & 92.66 & 100\\
        lefthyalinepaperfrontpants & PO,TO & 94.44 & 99.16\\
        leftmirrorside & HI & 93.58 & 96.93\\
        \bottomrule
    \end{tabular}
    }
    \label{table:sup1}
\end{table}

\begin{figure}[h]
    \centering
    \includegraphics[width=1\linewidth]{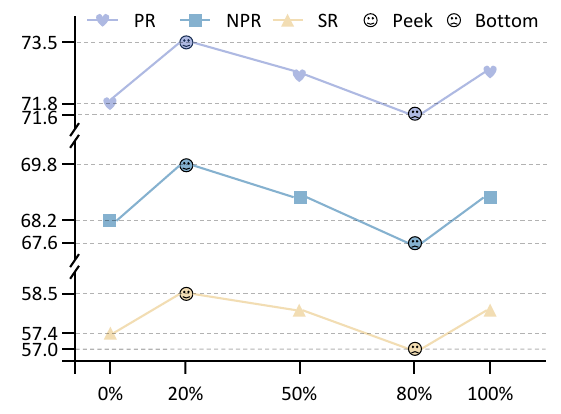}
    \caption{\textbf{Analysis of the \(\alpha\) ratio} in the Fourier visual prompts ablation experiment. We select the ratio at the peak as the final setting for the model's \(\alpha\). It is recommended to view in color for the best experience.}
    \label{fig:five}
\end{figure}

\textbf{Analysis of the \(\alpha\) ratio between visual prompts and Fourier prompts.} 
As depicted in Fig. \ref{fig:five}, we analyzed the different performances of visual prompts and Fourier prompts at various ratios in the benchmark in VFPTrack. \(\alpha\) represents the proportion of the Fourier prompts in the total prompts. The performance is not optimal when \(\alpha\) is 0\% or 100\%, meaning when only visual prompts or only Fourier prompts are used. Visual prompts carry spatial information that is intuitive for the model to understand the object's appearance and location in the original image space. However, they lack the frequency-domain information that can provide insights into the underlying patterns and global characteristics of the data. On the other hand, Fourier prompts offer a different perspective from the frequency domain but might lose some of the fine-grained spatial details. Only by setting an appropriate \(\alpha\) ratio between the visual and Fourier prompts can both the spatial and frequency domains be effectively combined. This combination enables the model to leverage the advantages of both types of prompts. For example, in a challenging tracking task where the object undergoes significant appearance changes, the visual prompts can help in quickly identifying the object's initial location based on spatial appearance, while the Fourier prompts can assist in maintaining the tracking when the object's appearance becomes distorted by providing frequency-domain stability. Thus, an appropriate \(\alpha\) ratio leads to the best performance.

\begin{figure*}[h]
    \centering
    \includegraphics[width=1\linewidth]{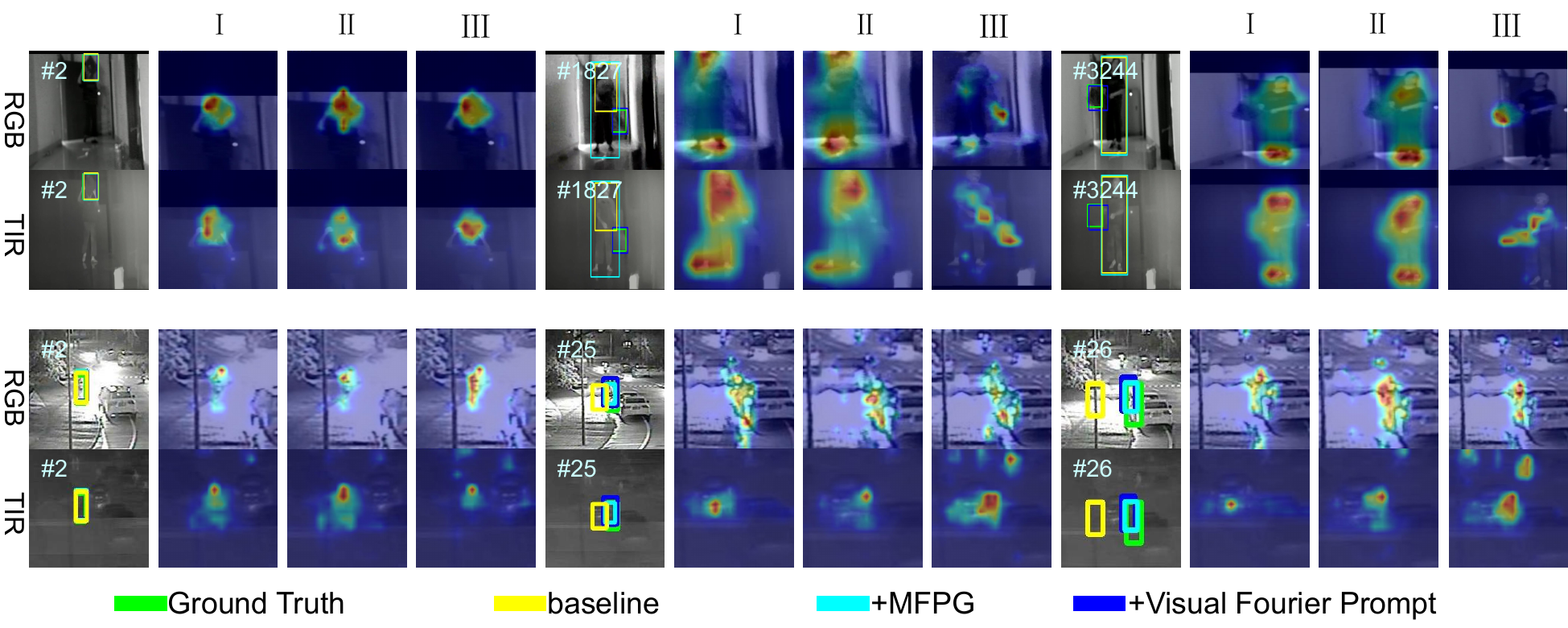}
    \caption{\textbf{Comparison} of attention maps in the search regions of the RGB and TIR modalities is visualized for two sequences from the LasHeR \cite{lasher}. The figure includes: Original frames with the tracking results highlighted, attention maps in the baseline tracker’s search regions (in Tab. I), attention maps after adding the MFPG module in the baseline tracker’s search regions(in Tab. II), and attention maps in the VFPTrack’s search regions (in Tab. III).}
    \label{fig:eight}
\end{figure*} 

\begin{figure}[!t]
    \centering
    \includegraphics[width=1\linewidth]{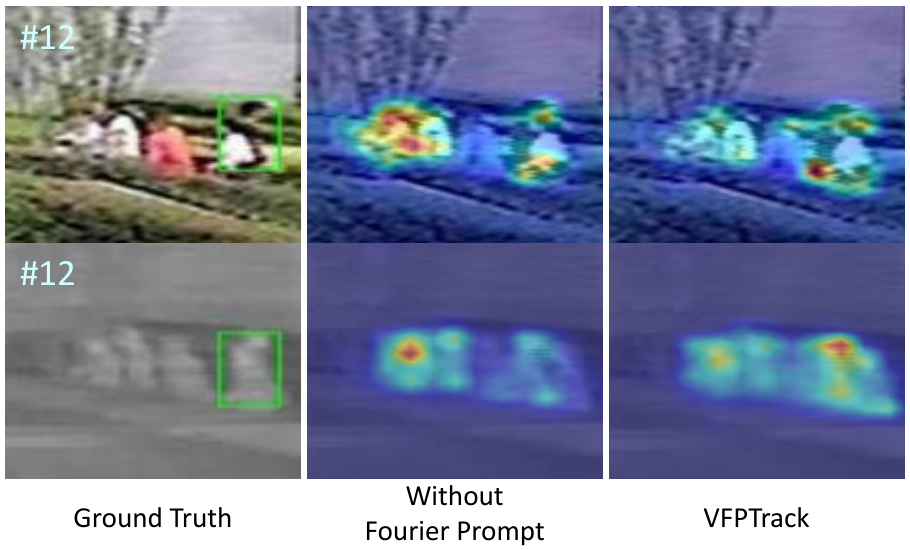}
    \caption{\textbf{Comparisons} of attention maps on sequence "blackdown" of LasHeR \cite{lasher}. The ground truth represents the original frames of the highlighted tracking object, "without Fourier prompt" indicates the attention map of the tracker after the removal of the Fourier prompt, and "VFPTrack" refers to the final attention map of our tracker.}
    \label{fig:six}
\end{figure} 

\begin{table}[h]
    \centering
    \caption{more quantitative comparison of the PR/SR scores of the visual Fourier prompt is made based on the nine attributes in the LasHeR \cite{lasher} dataset.}
    \resizebox{\linewidth}{!}{
    \begin{tabular}{c|c|c|c}
    \midrule
    \textbf{Attribute} & \textbf{w/o Visual Fourier Prompt} & \textbf{w/o Fourier Prompt} & \textbf{VFPTrack}\\
        \midrule 
        TO & 61.0/48.6 & 63.8/50.6 & \textbf{66.2}/\textbf{52.1}\\
        PO & 67.7/53.8 & 69.2/55.1 & \textbf{71.1}/\textbf{56.5}\\
        LI & 56.2/46.1 & 59.6/48.4 & \textbf{60.5}/\textbf{49.4}\\
        CM & 66.9/53.0 & 68.6/54.3 & \textbf{70.9}/\textbf{56.1}\\
        LR & 62.5/46.2 & 64.0/47.4 & \textbf{67.4}/\textbf{49.8}\\
        TC & 62.4/49.5 & 64.8/51.5 & \textbf{66.1}/\textbf{52.3}\\
        SA & 63.8/51.1 & 64.8/52.1 & \textbf{66.9}/\textbf{53.4}\\
        BC & 68.1/54.0 & 70.3/55.5 & \textbf{73.3}/\textbf{57.9}\\
        DEF & 75.7/61.0 & 78.0/62.8 & \textbf{78.7}/\textbf{62.9}\\
        \bottomrule
    \end{tabular}
    }
    \label{table:seven}
\end{table}

\textbf{Further Analysis of the Scene-Adaptive \(\alpha\) Ratio.}
In our experiments, the best overall performance was achieved when the proportion of frequency-domain prompts $\alpha$ was set to 20\%. From an information-theoretic perspective, spatial and frequency-domain prompts provide complementary cues: spatial prompts capture global structural and positional information, while frequency-domain prompts emphasize fine-grained details such as edges and textures. The weight $\alpha$ regulates the incorporation of high-frequency information; an excessively large value biases the representation toward redundant details, whereas an overly small value underutilizes frequency cues. The observation that \(\alpha=20\%\) yields the best performance can thus be interpreted as a balance point that maximizes the mutual information between spatial and frequency cues. To further examine this property, a scene-adaptive adjustment strategy was introduced, where RGB and TIR inputs were passed through a lightweight module composed of a linear layer followed by a non-linear activation to generate adaptive $\alpha$ values for each modality. On the LaSHeR dataset, this adaptive strategy produced results comparable to the fixed setting (within 1\% SR difference), and the analysis of representative sequences summarized in \ref{table:sup1} shows that $\alpha=20\%$ consistently dominates across challenging attributes such as occlusion, background clutter, and fast motion. These findings align with the fixed-$\alpha$ experiments and support the robustness of this ratio, while suggesting that more sophisticated adaptive strategies could be a promising direction for future research. It can be observed that the proportion corresponding to $\alpha=20\%$ consistently dominates across different challenging scenarios (see Tab. \ref{table:sup1}), which is consistent with the fixed $\alpha$ experiments and further supports the choice of this ratio.

\textbf{Analyze the challenges of visual Fourier prompts under different attributes.} As shown in Tab. \ref{table:seven}, "w/o Visual Fourier Prompt" represents the absence of the Visual Fourier Prompt component, while "w/o Fourier Prompt" indicates that only the visual prompt is used. These prompts are combined with image features, enabling the model to learn a joint representation that integrates both spatial and frequency information. Targets often exhibit distinctive frequency characteristics, such as high-frequency components associated with edges, textures, and fine-grained structures, whereas backgrounds typically contain smoother, low-frequency components. By incorporating frequency-domain prompts, the model highlights these differences, complementing spatial cues and enhancing feature extraction. Prior studies \cite{1, zeng2024visual, 2} have shown that frequency-based representations improve discrimination between objects and cluttered backgrounds. Consistent with these findings, our analyses in Tab. \ref{table:seven} and Fig. \ref{fig:six} demonstrate that Fourier prompts help distinguish frequency patterns between the target and background, resulting in more reliable tracking in challenging scenarios. Notably, our method shows significant improvements under Partial Occlusion (PO) and Total Occlusion (TO), indicating that combining visual and Fourier prompts effectively leverages complementary domain information. For attributes with fewer spatial cues, such as Low Resolution (LR) and Background Clutter (BC), adding Fourier prompts improves PR scores by over 3.0\% and SR scores by over 2.4\%, further highlighting the importance of frequency-domain information in RGB-T tracking. The attention associated with the Fourier prompt, particularly in occluded or low-resolution scenarios, is visualized in Fig. \ref{fig:six}, showing how frequency-domain features help the model focus on the target even when spatial cues are limited.

\begin{table}[h]
    \centering
    \caption{Conduct more quantitative comparison of the PR/SR scores based on different attributes in the LasHeR \cite{lasher} dataset.}
    \resizebox{\linewidth}{!}{
    \begin{tabular}{c|c|c|c|c}
    \midrule
    Attribute & mfDiMP & ViPT & BAT & VFPTrack\\
        \midrule 
        LI & 29.6/23.8 & 49.8/41.2 & 60.4/48.2 & \textbf{60.5}/\textbf{49.4}\\
        MB & 37.6/28.7 & 57.5/46.0 & 62.4/49.6 & \textbf{66.5}/\textbf{52.6}\\
        HO & 19.8/23.8 & 46.9/43.4 & 56.5/\textbf{51.0}& \textbf{58.2}/\textbf{51.0}\\
        TO & 32.2/25.0 & 57.7/46.2 & 64.1/51.1 & \textbf{66.2}/\textbf{52.1}\\
        PO & 39.7/30.8 & 62.4/50.4 & 67.5/54.0 & \textbf{71.1}/\textbf{56.5}\\
        NO & 76.5/57.5 & 84.1/68.5 & \textbf{90.2}/\textbf{73.3} & 89.5/72.9\\
        ARC & 37.8/30.9 & 59.4/49.5 & 63.1/51.9 & \textbf{66.2}/\textbf{54.1}\\
        SV & 45.2/34.9 & 65.0/52.5 & 69.7/55.9 & \textbf{73.1}/\textbf{58.4}\\
        FM & 41.3/32.4 & 63.2/51.5 & 68.5/55.1 & \textbf{71.9}/\textbf{57.8}\\
        OV & 40.6/34.9 & 76.2/65.0 & \textbf{76.8}/\textbf{66.0} & 72.3/62.3\\
        FL & 32.3/25.7 & 59.7/46.9 & 62.0/49.0 & \textbf{65.9}/\textbf{51.3}\\
        TC & 38.0/28.8 & 57.4/46.0 & 62.7/50.1 & \textbf{66.1}/\textbf{52.3}\\
        CM & 40.8/30.6 & 62.0/50.0 & 68.0/54.4 & \textbf{70.9}/\textbf{56.1}\\
        SA & 37.2/29.5 & 57.4/46.5 & 61.3/49.2 & \textbf{66.9}/\textbf{53.4}\\
        BC & 34.9/27.0 & 65.0/51.9 & 67.7/53.9 & \textbf{73.3}/\textbf{57.9}\\
        DEF & 40.3/34.2 & 67.6/55.8 & 72.2/58.5 & \textbf{78.7}/\textbf{62.9}\\
        LR & 40.2/25.6 & 56.7/41.8 & 62.7/46.2 & \textbf{67.4}/\textbf{49.8}\\
        AIV & 16.6/16.4 & 36.3/34.2 & 51.4/45.3 & \textbf{51.7}/\textbf{45.5}\\
        HI & 46.7/35.1 & 67.8/54.2 & 75.3/59.6 & \textbf{77.0}/\textbf{60.9}\\
        \bottomrule
    \end{tabular}
    }
    \label{table:eight}
\end{table}

\subsection{Visualization Analysis}
\textbf{Comparison across Various Extreme Attributes.} We compare VFPTrack with multiple trackers under 19 different attributes in the LasHeR dataset. The experimental results of its PR and SR scores are shown in Tab. \ref{table:eight}. Our model outperforms the others in the vast majority of extreme attributes. Notably, we achieve over 4\% higher SR scores under the "SA" and "BC" attributes, and the PR score under the "DEF" attribute exceeds 6\%. This further validates that our proposed model effectively leverages different domain information and modalities in complex environments to achieve superior multi-modal tracking performance.

\begin{figure}[t]
    \centering
    \includegraphics[width=1\linewidth]{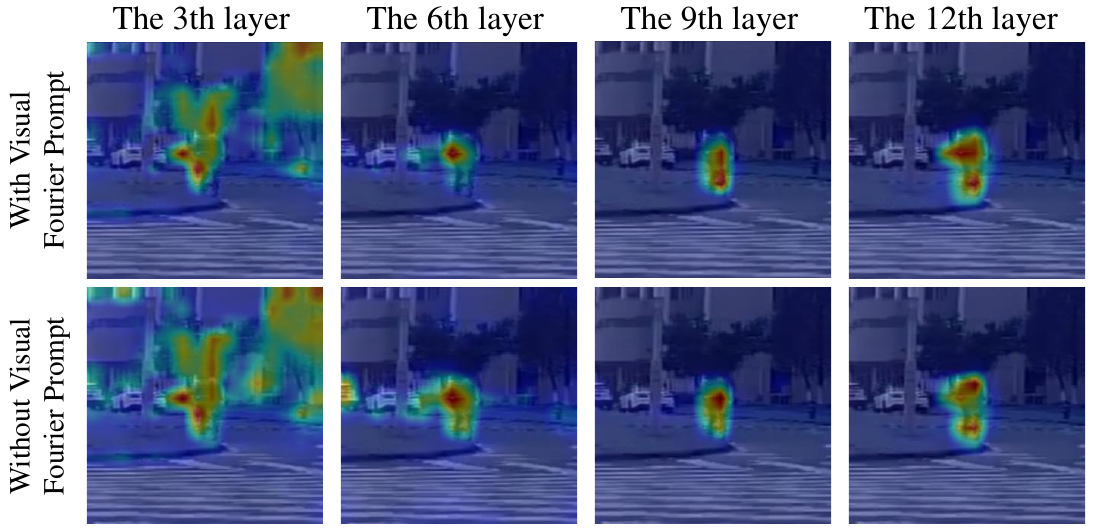}
    \caption{\textbf{Comparison} of attention maps at the 3rd, 6th, 9th, and 12th layers for a bike frame, without and with Visual Fourier Prompt tokens.}
    \label{fig:1-}
\end{figure} 

\begin{figure}[t]
    \centering
    \includegraphics[width=1\linewidth]{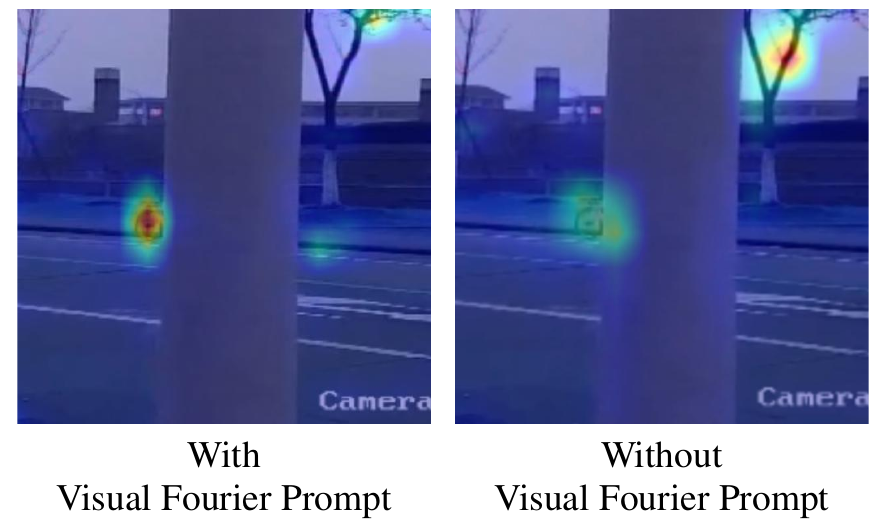}
    \caption{\textbf{Comparison} of 9th-layer attention maps under occlusion-and-reappearance, without and with Visual Fourier Prompt tokens. Red regions indicate stronger attention.}
    \label{fig:2-}
\end{figure}

\textbf{Visualization Results.} To more intuitively demonstrate the effectiveness of our proposed VFPTrack in RGB-T tracking scenarios, we visualize the comparison results with three of the most representative trackers from previous works. The results are shown in Fig. \ref{fig:seven}. Our method, by combining the advantages of frequency-domain and spatial-domain prompts, exhibits superior performance in scenarios involving occlusions and long-duration tracking (over 1K frames). Moreover, to validate the effectiveness of the proposed network, we conduct further attention visualization ablation experiments on different network components. The attention visualization results are shown in Fig. \ref{fig:eight}, where I represents the attention visualization of the baseline, and II and III correspond to the addition of the MFPG network component and the final VFPTrack we propose, respectively. From the first image sequence, it can be observed that as the tracking continues for an extended period, the image features are influenced by the continuous interaction between the two modalities, causing the tracked object to shift from a "bolster shaking" to a "person." However, under the combined effect of visual prompts and Fourier prompts, the integration of different domain information perspectives helps avoid this issue. Additionally, the second image sequence validates that our MFPG fully leverages the features from both modalities, ensuring that even in harsh environments with strong light interference, tracking does not experience the drift phenomenon observed in the baseline.

\textbf{Analysis of Layer-wise Effects of Visual Fourier Prompts on Feature Hierarchies.}
To examine the role of prompts across different layers, we visualize the attention maps at the 3rd, 6th, 9th, and 12th layers of the Transformer using a representative frame from the bike sequence. The original search region image and the visualization results are jointly shown in Fig. \ref{fig:1-}. In these maps, red regions indicate stronger attention responses.
At the 3rd layer, which is sensitive to low-level textures and fine-grained visual details, the model distributes attention broadly across informative regions. With prompts, the attention becomes less dispersed and more target-oriented. At the 6th layer, this effect is more pronounced, as attention with prompts is concentrated on the target, while the baseline still attends to the target but also allocates attention to irrelevant background regions. At the 9th and 12th layers, prompts enhance semantic discrimination, guiding the model to consistently highlight the target with stronger attention responses.
To further illustrate the role of prompts in deeper layers under challenging conditions, Fig. \ref{fig:2-} presents the 9th-layer attention maps for an occlusion-and-reappearance case. Prompts clearly improve semantic discrimination, allowing the model to maintain focus on the bicycle even when only partial features are visible, whereas the baseline still allocates considerable attention to irrelevant background objects.

\section{CONCLUSION}
In this work, we propose VFPTrack for RGB-T tracking. By introducing visual Fourier prompts, it enables the foundation model to integrate spatial and frequency domain information, thereby achieving more effective feature extraction. The proposed MFPG module generates simple and effective fused modality prompts. It facilitates the integration of modality features and establishes a complementary relationship between them. Extensive experiments on three RGB-T tracking benchmarks demonstrate the outstanding performance of VFPTrack.

\section*{Acknowledgments}
This work is supported by the Project of Guangxi Science and Technology (No.2024GXNSFGA010001 and 2025GXNSFAA069676), the National Natural Science Foundation of China (No.U23A20383, 62472109 and 62466051), the Guangxi "Young Bagui Scholar” Teams for Innovation and Research Project, the Research Project of Guangxi Normal University (No.2025DF001), the Innovation Project of Guangxi Graduate Education (XYCS2025123), and Guangxi Engineering Research Center of Educational Intelligent Technology.

\bibliographystyle{IEEEtran}

\bibliography{mian}

\newpage

\section{Appendix}
\textbf{Additional Backgrounds.} 
With the advancement of deep learning techniques \cite{gong2021eliminate,gong2022person,gong2024cross,peng2025directing,peng2024lightweight,peng2025boosting,peng2025towards,lu2024mace,lu2023tf,lu2024robust,li2025set,gao2024eraseanything,ren2025all,liu2025shifting,liu2025global,liu2025video,liu2025m2ist,Leo2024AgentNet,Zhang2025SensitivityLoRA,Zhang2025TimeLLaMA,Wang2025OneImage,Qiu2025IntentVCNet,Leo2024ICAICA,Leo2024MedDoc,Sun2024RadiologyLLM,gsq,zhou2024lidarptq,zhou2024information,pillarhist,zhou2023fastpillars,he2024diffusion,he2025segment,he2023hqg,he2025unfoldir,he2025run,he2025reti,he2024weakly,xiao2024survey,he2023strategic,he2023camouflaged,he2023degradation} and the potential to eliminate the need for large-scale labeled data, self-supervised tracking has attracted increasing attention from researchers. Taking advantage of intrinsic correlations in unlabeled video data, such as temporal consistency, self-supervised tracking has shown promising results in relatively simple tracking scenarios. However, in long-term complex unlabeled tracking settings, it remains a significant challenge to capture cross-frame motion patterns and to learn robust target representations.

\textbf{Evaluation Metrics.} The tracking performance is evaluated using the toolkit corresponding to the dataset. We follow the evaluation protocol of published datasets and employ three metrics to ensure a fair comparison across various tracking methods, including success score (AUC), normalized precision score (P$_{\rm{Norm}}$), and precision score (P).

\end{document}